\documentclass[conference]{IEEEtran}
\usepackage{times}
\usepackage{epsfig}
\usepackage{graphicx}
\usepackage{amsmath}
\usepackage{amssymb}
\usepackage{cite}
\usepackage{fancyhdr}
\usepackage{hyperref}

\fancypagestyle{firstpage}{
  \fancyhf{}
  
  \fancyfoot[C]{
    \footnotesize{
      © 2024 IEEE. Personal use of this material is permitted. Permission from IEEE must be obtained for all other uses, in any current or future media, including reprinting/republishing this material for advertising or promotional purposes, creating new collective works, for resale or redistribution to servers or lists, or reuse of any copyrighted component of this work in other works.
    }
  }
}

\fancypagestyle{otherpages}{
  \fancyhf{}
  
  \fancyfoot[C]{
    \begin{center}
      This paper has been accepted in IJCB 2024 Special Session, Generative AI for Futuristic Biometrics.
    \end{center}
  }
}

\pdfoutput=1

\begin{document}

\title{A Secure and Private Ensemble Matcher Using Multi-Vault Obfuscated Templates}

\author{
Babak Poorebrahim Gilkalaye \quad Shubhabrata Mukherjee \quad Reza Derakhshani\\
School of Science and Engineering, University of Missouri-Kansas City\\
{\tt\small bpktk@mail.umkc.edu \quad smpw5@umsystem.edu \quad derakhshanir@umkc.edu}
}

\maketitle
\thispagestyle{firstpage} 

\pagestyle{otherpages}

\begin{abstract}
Generative AI has revolutionized modern machine learning by providing unprecedented realism, diversity, and efficiency in data generation. This technology holds immense potential for biometrics, including for securing sensitive and personally identifiable information. Given the irrevocability of biometric samples and mounting privacy concerns, biometric template security and secure matching are among the most sought-after features of modern biometric systems. This paper proposes a novel obfuscation method using Generative AI to enhance biometric template security. Our approach utilizes synthetic facial images generated by a Generative Adversarial Network (GAN) as ``random chaff points" within a secure vault system. Our method creates $n$ sub-templates from the original template, each obfuscated with $m$ GAN chaff points. During verification, $s$ closest vectors to the biometric query are retrieved from each vault and combined to generate hash values, which are then compared with the stored hash value. Thus, our method safeguards user identities during the training and deployment phases by employing the GAN-generated synthetic images. Our protocol was tested using the AT\&T, GT, and LFW face datasets, achieving ROC areas under the curve of 0.99, 0.99, and 0.90, respectively. Our results demonstrate that the proposed method can maintain high accuracy and reasonable computational complexity comparable to those unprotected template methods while significantly enhancing security and privacy, underscoring the potential of Generative AI in developing proactive defensive strategies for biometric systems.
\end{abstract}

\section{Introduction}
Biometric user verification has become ubiquitous, but its vulnerabilities to various attacks and privacy breaches emphasize the need for enhanced security measures. Recent advances in Generative AI (GenAI) have transformed the field of machine learning, offering new opportunities for improving biometric security and privacy. GenAI's ability to generate highly realistic and diverse synthetic data has been leveraged in various applications, including biometric security~\cite{goodfellow2014generative, karras2020stylegan}. When it comes to biometric security and privacy, GenAI may be used to synthesize biometric data for template obfuscation thus safeguarding user privacy. Unlike traditional methods, GenAI-produced chaff points can be highly varied and realistic, making it exceedingly difficult for attackers to distinguish between genuine and synthetic data, significantly enhancing biometric template security~\cite{yang2022genai, wang2023genai, chen2022genai}.

Biometric systems have been vulnerable to various attack vectors, including sensor hijacks, injection of forged biometric templates, and network attacks on servers~\cite{8590812}. Such vulnerabilities may expose biometric systems to correlation attacks~\cite{5337551}, presentation attacks, replay attacks, Man-in-the-Middle attacks, Denial-of-Service attacks, or Brute-Force attacks~\cite{husseis2019survey, 8506419, 4105331, 7029643}. Security incidents and privacy breaches through social media platforms, healthcare data, and government services~\cite{ur2019facebook, segun2017healthcare, singh2021aadhaar} exemplify these vulnerabilities. The need for securing biomedical and biometric reference data is highlighted by the substantial increase in data breaches across US institutions and hospitals, including data theft, unauthorized access, and hacking, as reported by the HIPAA journal~\cite{HIPAAJournal2024}.

To protect stored biometric references from such attacks, researchers have proposed various security schemes, such as random projection, feature disentanglement, fuzzy vaults with auxiliary (helper) data, and deep learning-based concealable multi-biometrics to mitigate the earlier-mentioned privacy and security issues~\cite{5382573, 9320295, 1640610, abdellatef2020cancelable}. However, many of these approaches are either computationally too expensive or not completely secure against certain attack vectors. In this work, we propose an encrypted vault approach to securely store biometric templates. Our approach uses mutually exclusive embeddings generated by various deep learning models, combined and then securely encrypted using SHA-512-based encryption. This technique incorporates 2000-4000 AI-generated face embeddings as chaff points, stored alongside the genuine template in a secure vault. This makes it computationally infeasible to distinguish the original template from synthetic data. This attribute makes our approach robust against brute-force attacks.

The reference feature vector, derived from the enrollment image and used for comparison with other incoming images, is called an enrollment template. For example, in the context of a Convolutional Neural Network (CNN) trained for face matching, this could be the flattened input to an FC layer. If the target image corresponds to the same person or object, for instance by the corresponding templates having a cosine similarity below a set threshold, the claimant is accepted. In this case, the claimant is genuine, and the acceptance is a true positive or genuine accept. If the claimant is an impostor (the incoming image is not from the same identity) and they manage to pass the template comparison, the result is a false or impostor accept. A genuine comparison is anticipated to yield a higher similarity, indicating that the feature vectors derived from the images closely match.

\begin{figure}[htbp]
\centerline{\resizebox{0.54\textwidth}{!}{\includegraphics{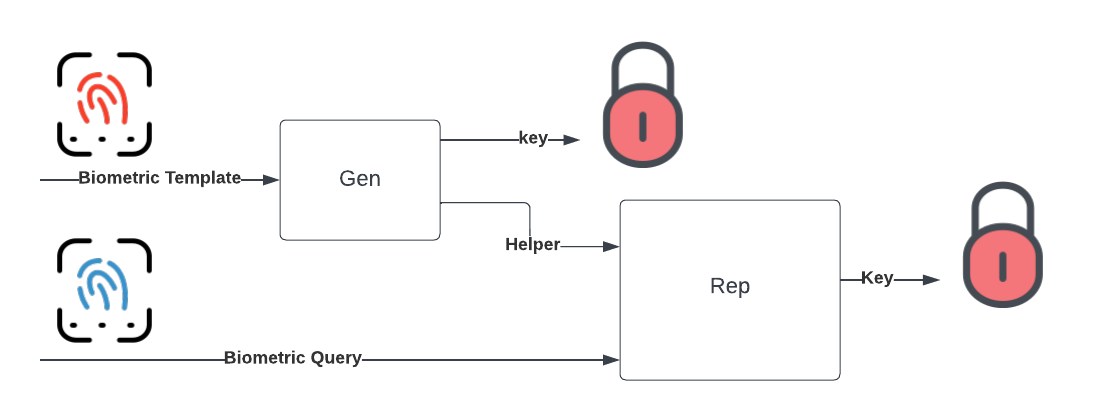}}}
\caption{Fuzzy extractor scheme}
\label{fuzzy_sketch}
\vspace{-10pt} 
\end{figure}

Fig.~\ref{fuzzy_sketch} describes a fuzzy extractor scheme and its terminology. We will use the same structure and terminology for our proposed obfuscation method. The \textsc{Gen} function generates a secret key and auxiliary data $P$ from a biometric template $t$. The \textsc{Rep} function takes a biometric query $q$ and the auxiliary data as its input. If dist($q,t$) is smaller than a preset threshold, then it generates the correct key. In other words, the auxiliary data $P$ helps to remove noise from query $q$ in order to generate $t$. The auxiliary data $P$ is required to be \textit{secure} and reveal limited information about template $t$ to preserve the privacy of template $t$. To satisfy the information-theoretic security, the entropy loss must be at most $\mathcal{L}$, as further detailed below~\cite{dodis2004fuzzy}, \cite{dodis2008fuzzy}, \cite{chang2006hiding}.

The rest of this paper is organized as follows: Section II reviews related work. Section III describes our methodology. Section IV describes the security capability of our scheme. Section V outlines the experimental setup. Section VI presents our results and their implications. Finally, we conclude with a summary of our method's unique attributes, challenges faced, and future work.

\section{Related Work}

Many methods have been introduced over the years to preserve the privacy of biometric data. For instance, \cite{1377174} used a de-identification technique, but anonymity was not guaranteed. Other methods explored encryption-based schemes \cite{7420576}, but these are vulnerable to various attacks, including those that exploit high False Acceptance Rates (FAR), where an attacker attempts to gain unauthorized access by presenting multiple fraudulent biometric samples until one is falsely accepted as genuine. Our approach, demonstrably secure against such attacks (Section IV), offers significant improvements. Techniques utilizing Convolutional Neural Networks (CNNs) have been used to protect face templates while maintaining good matching performance \cite{8575550}. Blockchain-based storage of biometric templates has also been discussed \cite{9025645}, but it often incurs high execution times. Adversarial learning for concealing information within a learned space has also been explored \cite{9163294}, focusing on disentangling identity from attribute information to mitigate soft-biometric information in face templates \cite{9469725, 9320295}. Protecting source data by combining biometric modalities (e.g., fingerprint and iris scans) using various fusion techniques has been studied \cite{singh2023dual, 7062734}. While enhancing security through redundancy, it introduces a single point of failure, making the entire system vulnerable if one trait is compromised. It is also possible to show the unlinkability of the protected templates using different existing measures such as maximal leakage~\cite{shahreza2023measuring}.
In recent years, Generative AI (GenAI) has gained traction in biometric privacy. GenAI's ability to generate highly realistic synthetic biometric data shows promise for enhancing privacy and security. For instance, \cite{Fang2022} and \cite{Liu2023} demonstrated the effectiveness of Generative Adversarial Networks (GANs) in creating synthetic facial images that can obfuscate original data, and thus protect user identities. Similarly, \cite{Yang2022} explored using GenAI to generate diverse biometric data, improving the robustness of biometric systems against various attacks.
Recent studies have investigated GenAI's role in creating synthetic multimodal biometric data, for instance combining facial images with other modalities such as voice and fingerprints \cite{Wang2023, Chen2022}. This approach not only enhances security but also provides additional layers of privacy protection by making it more difficult to link synthetic data to real individuals. Furthermore, \cite{Zhang2022} highlights the potential of GenAI in generating synthetic data for training biometric systems, reducing reliance on real biometric data, and thus mitigating some privacy concerns. 

One can leverage existing measures, such as maximal leakage \cite{shahreza2023measuring}, to demonstrate the unlikability of protected templates. Our approach leverages such GenAI advances to provide a robust and secure method for protecting biometric templates, addressing a number of the limitations of previous methods while being computationally efficient.

\section{Methodology}
\subsection{Main Idea}
In this section, we will first informally explain the main idea behind the proposed method, and then provide the formalized algorithm. The idea is to divide the biometric template into $m$ sub-templates and hide them with $n$ chaff points to make it computationally impractical for an adversary to find the template. 
One solution would be to hide a template $t$ with $n$ chaff points in a public vault, $V=[r_1, ..., t, ..., r_n]$, and store $f(t)$, where $f$ is a one-way function such as SHA-256. Now the question is how hard it would be for an adversary to find the index of $t$ in $V$. To make it computationally impractical for the adversary to find $t$, $n$ must be a very big number $ (\lambda$ (security parameter)), as the best strategy to find the template $t$ is to compute $f(x)$ for all $x$ in the vault to find the template $t$. However, this method is also computationally impractical for the legitimate user. By way of example, assume we will have $f(t)$ and one vault containing $1 + 2^{80}$ vectors as auxiliary data. We assume the inserted chaff points are indistinguishable from the template $t$ in the collection. By indistinguishable chaff points, we mean that the chaff points represent realistic face templates that an attacker cannot differentiate from the actual template. We achieve this using StyleGAN2 to generate synthetic face embeddings that closely mimic the statistical properties of genuine templates, making it extremely challenging to identify the original among them. For matching a query $q$, we compute the distance of $q$ from all points inside the vault and retrieve $m$ (depending on the threshold of the system) closest points to $q$: ${p_1,p_2,...,p_m}$, and compute $f(p_1), f(p_2),..., f(p_m)$ and check if there are any matches between these values and $f(t)$. This system is secure under certain conditions, but impractical because during the authentication the user needs to compute the desired distance of the query $q$ vs. $1 + 2^{80}$ vectors. 

Thus we propose generating $n$ sub-templates $t_1, t_2,..., t_n$ from template $t$, using protocol $\mathcal{G}$, such that $\mathcal{G}(t): t_1,t_2,...,t_n$ and hide each of them with $m$ chaff points. In the end, we store $f([t_1,t_2,...,t_n])$ (, where $f$ is a one-way function), in $n$ vaults such that each vault has $m+1$ vectors. To make it computationally impractical for the adversary to find the template by brute force, we need $m^n$ to be very large. Unlike the earlier impractical example, we can choose $m$ and $n$ to garner computational security against brute force attacks while having a practical solution. In the forthcoming experimental demonstration, we use $m=2000$, and $n=5$, so that $m^n = 2^{54.82}$ to get acceptable security against a brute force attack. When matching query $q$, we generate $q_1,q_2,...q_n$ using protocol $\mathcal{G}(q): q_1,q_2,...,q_n$ (protocol $\mathcal{G}$ will be explained later). The distances between $q_i$ and all $m+1$ vectors in the corresponding vault are then computed and the top $s$ vectors from each vault are retrieved. The threshold of the system is $s$ and the distance can be computed by any distance metric, such as cosine similarity. At this point, we have retrieved $sn$ vectors. Next, we construct all $s^n$ possible combinations of vectors $v_1,v_2,...v_{s^n}$ and compute $f(v_1),f(v_2),...f(v_{s^n})$ and check if there are any matches between the stored value of $f([t_1,t_2,...,t_n])$ and the $f(v_i)$. If $q$ is a genuine vector, it should be able to retrieve $t_1, t_2, ..., t_n$ with the threshold~$s$. If $q$ is an imposter, it will fail to retrieve $t_1,t_2,...t_n$ from the vaults. It should be noted that $s$ cannot be a large number as it would render the biometric matching computationally impractical. 

\subsection{True Positive Rate (TPR) Improvement}

It can be seen that a query needs to satisfy all $n$ vaults in order to succeed. Thus we relax this condition and build the system such that a query can succeed if it satisfies $k$ out of $n$ vaults. To do so, after the generation of sub-templates $t_1,t_2,...t_n$, we need to store all $n \choose k$ possible combinations of $f(t_i)$. To satisfy the asymptotic security, we need to make sure $m^k$ is computationally impractical for a computationally bounded adversary, where $m$ is the number of chaff points per vault. 

\subsection{The Algorithm}
Our algorithm $\textit{Alg}$ consists of a \textsc{Gen} protocol, a \textsc{Rep} protocol, a key generation protocol $\mathcal{K}$, a sub-template Generator $\mathcal{G}$, a one-way function $f$, and a security parameter $\gamma$. We assume that the template $t$ is sampled from the distribution $\mathcal{D}$.

\textbf{Key Generation $\mathcal{K}(\gamma)$}: This protocol chooses $n$ (number of vaults) and $m$ (number of chaff points)  so that $m^n > 2 ^\gamma$. Then it generates $mn$ random samples from distribution $\mathcal{D}$ for $mn$ chaff points $K=c_{11},c_{12}...,c_{1m},c_{21},c_{22},c_{21},...c_{2m},...,c_{nm}$. Two types of random numbers $R_{1n}$ and $R_{2n}$ are generated. Where $R_{1n}$ are $n$ scalars and $R_{2n}$ are small vectors with the same dimension as the template. These will be used to hide the template in each vault further, as follows. 
Let $d$ be the cosine similarity of two vectors, then: 

$d(A,B)$ = $d(A,rB)$ for any arbitrary scalar $r$, and:

$d(A,B)$ $\approx$  $d(A,r_1B+R_1)$ for any arbitrary scalar $r_1$ and small vector $R_1$

\textbf{Sub-template Generator} $\mathcal{G}(t,n)$: This protocol takes template $t$ and the number of vaults $n$ as its inputs and generates $n$ independent sub-templates $T=(t_1,t_2,...,t_n)$. In the experimental section, we propose one solution for such a function.  

\textbf{\textsc{Gen}}: $\textsc{Gen}(T,K)$  This protocol places each sub-template $t_i$ and its corresponding chaff points $c_{i,j}$, for all $j={1,..,m}$ inside a vault in no specific order. It then stores $P_1 =f(t_1,t_2,...t_n)$ and outputs the helper $P$:

\textbf{\textsc{Rep}}: $\textsc{Rep}(C,q,tr)$: With $q$ representing a new biometric query using protocol $\mathcal{G}$, $Q = (q_1,q_2,,...q_n)$ is generated and the closest $tr$  vectors are retrieved from each vault. The threshold of the system is denoted by $tr$. Note that $tr$ is different from the original well-known face matching threshold.  $tr$ may be adjusted to increase or decrease the False Acceptance Rate (FAR). In other words, if $tr = 1$ we only retrieve the closest vector in each vault. This will drastically reduce the False Positive Rate (FPR), but simultaneously it reduces the TPR.  

The following are retrieved from the $n$ vaults: 

$\{w_{11},w_{12},...,w_{1tr}\}$, $\{w_{21},w_{22},...,w_{2tr}\}$...

$\{w_{n1},w_{n2},...,w_{ntr}\}$

All $n \choose tr$ possible combinations of these vectors are created and passed to the one-way function $f$. If there is any match between the generated hash values and the stored hash value $P_1$, then the query $q$ is authenticated and the protocol outputs $1$ (success), otherwise, it outputs $0$ (fail).

\begin{figure*}[hbt!]
\centerline{\includegraphics[width=.8\textwidth]{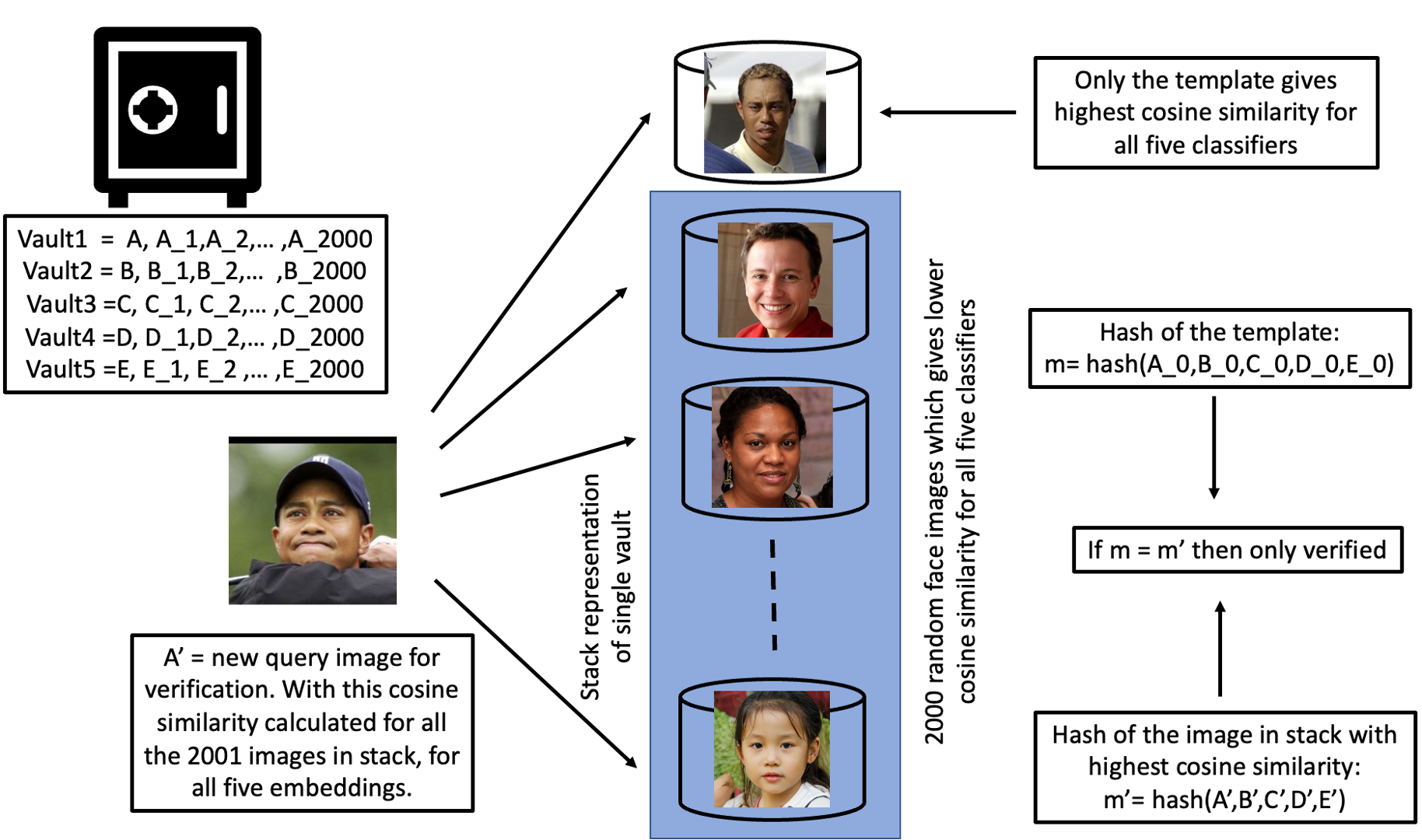}}
\caption{Embedding generated by the secure hashing verification}
\label{authn}
\end{figure*}

\section{Security }
An ideal biometric security scheme needs to satisfy \textbf{\textit{Revocability}}, \textbf{\textit{Diversity}}, \textbf{\textit{Accuracy}}, and \textbf{\textit{Security}} \cite{maltoni2009handbook} requirements. Our proposed scheme meets all these requirements, as detailed below.  

\textbf{\textit{Revocability}}: A biometric system needs to be revocable because if it is compromised, or if the user wishes to no longer be identified, they should be able to revoke their reference. Our scheme's templates are revocable by choosing a new set of chaff points and a new sub-template generator. This process is further facilitated by the ease of GAN-based chaff generation.   

\textbf{\textit{Diversity}}: A secure template must not allow cross-matching across databases, thereby ensuring the user's privacy. In our system, for a false positive to occur, the imposter must meet the authentication criteria of each vault and successfully retrieve all segments of the template, hence our system mitigates the cross-matching as compared to systems using one model. It also could be verified throughout the experimental section, the FPR is very small, because an imposter needs to satisfy multiple vaults to be able to positively match. 

\textbf{\textit{Accuracy}}: Our method shows minimum accuracy degradation with a 0.99 area under the curve (AUC) for both the AT\&T ~\cite{att_facedatabase} and Georgia Tech (GT) face databases~\cite{Nefian1999}. 

\textbf{\textit{Security}}: To show that our system is secure, we need to show that the auxiliary data $P=(P_1,P_2)$ is not revealing too much information and there is enough entropy in this system. Chaff points are generated from the same distribution that the corresponding sub-template is generated. This is achieved using StyleGAN2, which was originally trained on the Flickr-Faces-HQ (FFHQ)~\cite{karras2019style}dataset containing 70,000 high-quality, diverse human face images. This training ensures that the generated chaff points closely mimic the statistical properties and distribution of real face embeddings. Hence we can say the sub-templates inside each vault are indistinguishable from other chaff points there. Getting this indistinguishability between the sub-template and chaff points inside a vault is straightforward; the challenge is to achieve indistinguishability between sub-templates and chaff points. In other words, if an adversary knows that $t_1$ is a sub-template,  other $t_i$  should remain indistinguishable from chaff points, we achieve this by using independent models, given two models $M_1, M_2$, and template $t$ and a random face $r$, let $===$ be coming from the same image. 

$|Pr[M_1(t) \text{===} M_2(t)] - Pr[M_1(t) \text{===} M_2(r)]| = \epsilon$

To achieve this, we must make sure that the sub-templates are independent. If we assume everything looks completely random inside vaults, then we can claim the best strategy to find the template for any adversary is to check all possible combinations of elements in vaults, which we have made impractical by choosing proper $n$,$m$. As a result, our approach provides superior security against brute force attacks, as well as faster face acceptance and rejection capabilities.

It is clear that given a face template $t$, a random face $r$, and two biometric models $M1$ and $M2$, $y_1 = M1(t)$ and $y_2~=~M2(t)$ cannot be completely independent. However, in our case, we are using big neural networks. With this assumption and if $M1$ and $M2$ have completely different structures, it is not going to be easy to distinguish between $M1(t)$ and $M1(r)$ given  $M2(t)$. In other words, if we tell the adversary $M2(t)$ belongs to person $A$, and then if we ask him to tell us which one of $M1(t)$ and $M1(r)$ belongs to the same person, it will not be easy for the adversary to solve this problem. 

One approach is to reverse engineer the template to recreate it and see which one of reverse $M1(t)$ and reverse $M1(r)$ is closer to the reverse $M2(t)$. We have addressed this attack by adding randomness; now the adversary needs to distinguish between $R_2M1(t) + R_2'$ and $R_2M1(r) + R_2''$, given $R_1M2(t) + R_1'$.

\section{Experiments}
\subsection{Experimental Setup}
Per common practice in face recognition, we use cosine similarity for matching deep face templates~\cite{nguyen2010cosine}. Cosine similarity between any two vectors is given by:

\begin{equation}
\cos ({\bf A},{\bf B})= {{\bf A} {\bf B} \over \|{\bf A}\| \|{\bf B}\|} = \frac{ \sum_{i=1}^{n}{{\bf A}_i{\bf B}_i} }{ \sqrt{\sum_{i=1}^{n}{({\bf A}_i)^2}} \sqrt{\sum_{i=1}^{n}{({\bf B}_i)^2}} }
\end{equation}

Where A and B are the two feature vectors being compared.

\begin{figure}[hbt!]
\centerline{\includegraphics[width=0.8\columnwidth]{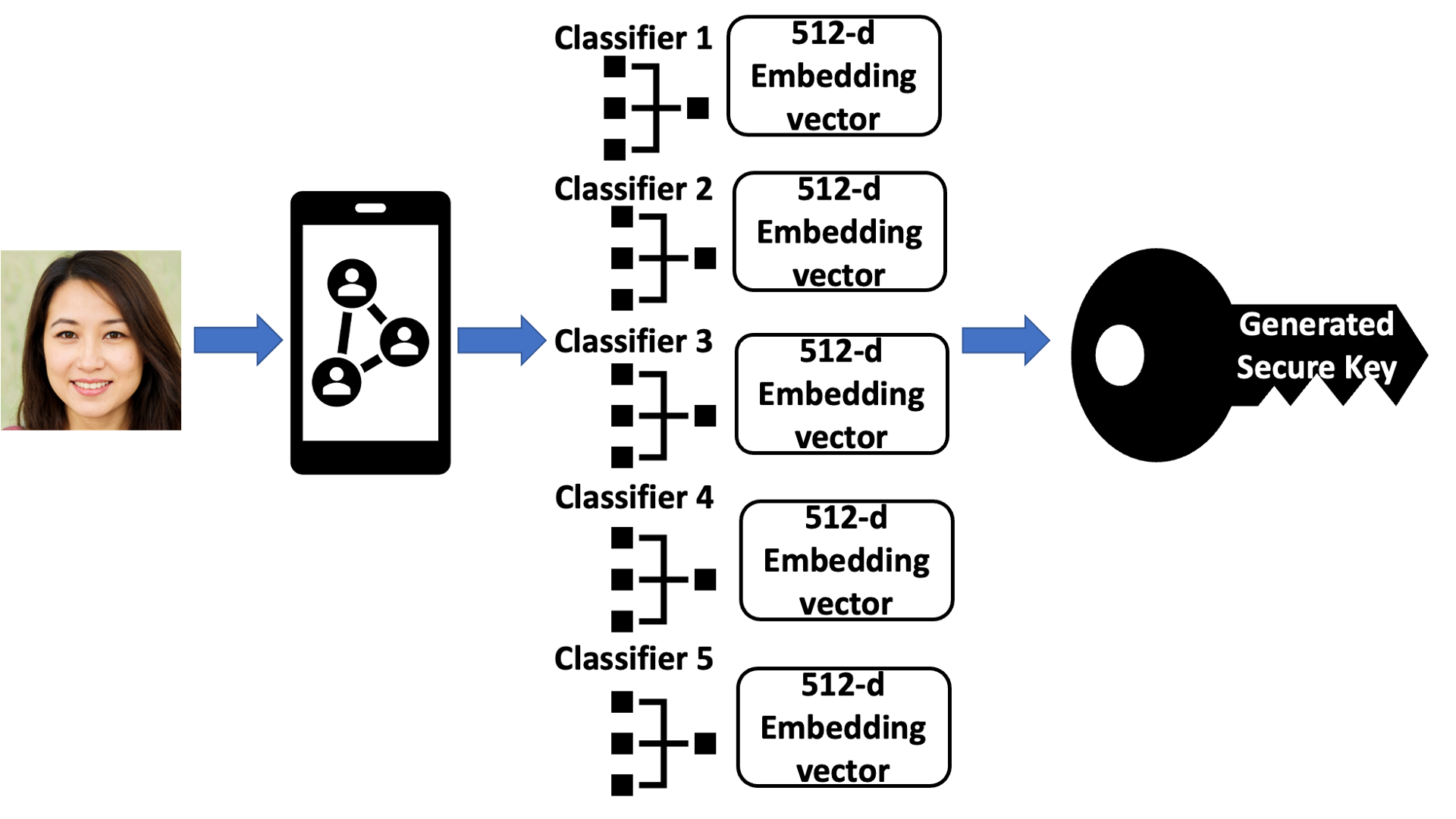}}
\caption{Generating key using five different classifiers}
\label{hashpic}
\end{figure}

In our research, a face image is transformed into an embedding using five different deep learning methods (Fig. \ref{hashpic}). Each of these embeddings is produced by a different convolutional neural network (CNN) model. The 512-dimensional feature vectors ($\boldsymbol{\mathit{e}}_1, \boldsymbol{\mathit{e}}_2, \boldsymbol{\mathit{e}}_3, \boldsymbol{\mathit{e}}_4, \boldsymbol{\mathit{e}}_5$) are numerically distinct, as verified by cosine similarity comparisons showing low similarity scores between vectors of different vaults. A total of five secure vaults is constructed. Each vault contains 2001 embeddings, of which 2000 embeddings are generated using 2000 random images of human faces and only 1 embedding is generated from an intended face image. We employed StyleGAN2 \cite{karras2019style} to generate realistic human face images for our chaff points. StyleGAN2 was chosen for its superior image quality, diversity, and photo-realism compared to alternatives like PGGAN \cite{karras2017progressive} and StarGAN \cite{choi2018stargan}. Its improved fine-scale detail, reduced artifacts, and disentangled latent space make it ideal for creating indistinguishable and diverse chaff points for our secure vault system. These images were then used to create random chaff points for the secure vault.\footnote{To achieve this, we conducted an iterative process, generating 2000 embeddings from 2000 realistic face images obtained from the website thispersondoesnotexist.com} This approach improves the privacy safeguards for the proposed method during the development and evaluation of the Multi-Vault Obfuscated Templates algorithm. The same structure, consisting of 2000 embeddings generated using random images and 1 embedding from an intended face image (i.e. the template), is replicated in all five vaults. To ensure security, the hash value of the five different embeddings $H = hash(e1,e2,e3,e4,e5)$ is stored.

\subsection{Verification Process}
The process for a single vault is shown in Fig.~\ref{authn}. For a biometric verification match:

\begin{itemize}
    \item Five distinct embeddings ($A'$, $B'$, $C'$, $D'$, $E'$) are generated using five separate classifiers based on the image of the person seeking verification (referred to as the query).
    \item Each resultant embedding is compared against all embeddings across the corresponding vault, resulting in the calculation of a total of 2001*5 cosine similarity scores.
    \item The embedding with the highest cosine similarity score is selected from each vault, leading to the choice of five winners, one from each vault.
    \item A secure 64-bit hash, denoted as $m'$, is constructed by combining the five embeddings selected in the previous step.
    \item The query is verified only if this hash matches the previously generated hash ( $H = hash(e1,e2,e3,e4,e5)$) of the template.
    However, in our implementation, we selected the top two or three embeddings from each vault and generated all possible $5^2$ or $5^3$ hashes, and compared them with the stored hash. This allows for relaxing the FRR of the system.
\end{itemize}

A higher number of random images can be used to increase security. A lower and higher number of classifiers can be used depending upon the scalability requirements. 

\subsection{Pre-trained Models and Datasets}
A set of five different pre-trained models have been used here to construct the end-to-end secure privacy framework.  The first model is InceptionResnetV1; this particular model~\cite{szegedy2017inception} has been pre-trained on the VGGFace2dataset.~\cite{cao2018vggface2,timesler}. As a second classifier, another variation of InceptionResnetV1 has been used, but this version was pre-trained with CASIA-Webface~\cite{yi2014learning}. For both models, one and two, MTCNN~\cite{zhang2016joint} has been used as a preprocessing for face cropping. More precisely, MTCNN cropped a 250x250 dimension face to improve the model accuracy. As a third and fourth classifier, two different versions of insightFace~\cite{deng2019arcface} have been used. The third pretrained classifier, named ``buffalo", is an implementation of the ResNet50 model pretrained on the WebFace600K dataset. This dataset contains 600,000 unique identities and is derived from the larger WebFace42M dataset~\cite{zhu2021webface260m} which contains 42 million identities. The fourth pre-trained model antelopev2 is an implementation of ResNet100 pretrained using Glint360K~\cite{an2021partial}, which contains 17091657 images of 360232 individuals. The fifth model is an arcface-based implementation of ResNet100~\cite{he2016deep}; for this version, some additional preprocessing has been used including image resizing, detection of faces using a face detector, and face alignment.

Our method was evaluated using three widely recognized face datasets: the AT\&T Database of Faces (AT\&T), Labeled Faces in the Wild (LFW), and Georgia Tech Face Database (GT). AT\&T contains 400 images of 40 subjects with controlled variations, LFW includes over 13,000 unconstrained face images collected from the web, and GT comprises 750 color images of 50 subjects with diverse facial expressions and illumination conditions. These datasets were selected to provide a comprehensive evaluation across different controlled and uncontrolled conditions, varying numbers of subjects, and diverse image characteristics, allowing us to assess our method's performance in a range of scenarios.

\section{Result and Analysis}

\begin{figure}[htbp]
\centerline{\includegraphics[width=.7\columnwidth]{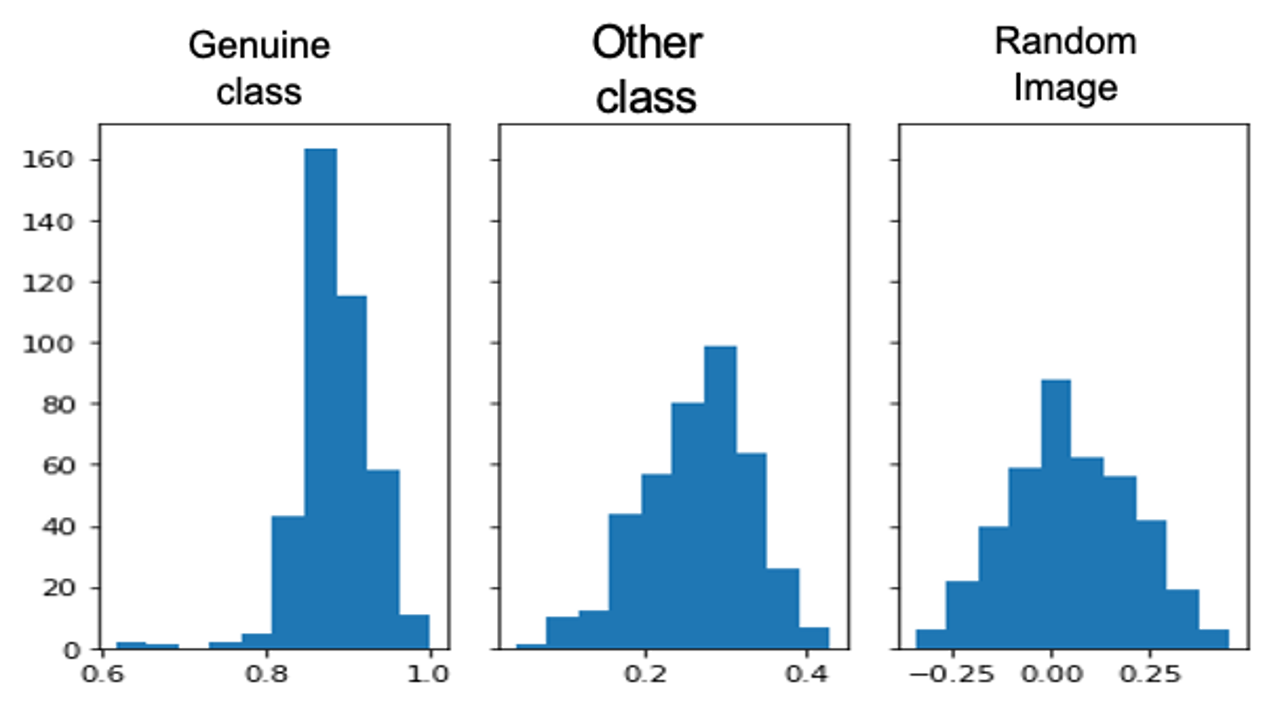}}
\caption{Distribution of Cosine Similarity for Genuine, Other, and Random Facial Images}
\label{dist1}
\end{figure}

Our results are mainly described from two perspectives, accuracy and time complexity performance. Fig.~\ref{dist1} shows the comparison between the distribution of cosine similarity scores for genuine and imposter face images. The leftmost distribution is for the genuine cosine similarities, which mostly lie between 0.8 to 1. The middle distribution is for the imposter cosine similarity computed from the target template and other (non-mated) class images from the \textit{same} dataset (e.g., same AT\&T dataset but different identities), with the score distribution mostly covering 0.2 to 0.4. The rightmost distribution also represents imposter cosine similarities but is computed by comparing the target templates and the random face images generated by GAN (as mentioned in the experimental setup), with the scores being mostly between -0.25 and 0.25. The constrained feature space of faces within a single dataset promotes higher cosine similarity (0.2-0.4) due to inherent structural similarities. Conversely, the inclusion of random external faces introduces greater vector variance, leading to a broader cosine similarity distribution (-0.25 to 0.25). These comparisons show how well the cosine similarity can distinguish the embeddings generated from face images. The corresponding ROC AUC of cosine similarity scores reinforces this fact.

\subsection{Individual Classifier Performance}

\begin{figure*}[hbt!]
\centerline{\includegraphics[width=.7\textwidth]{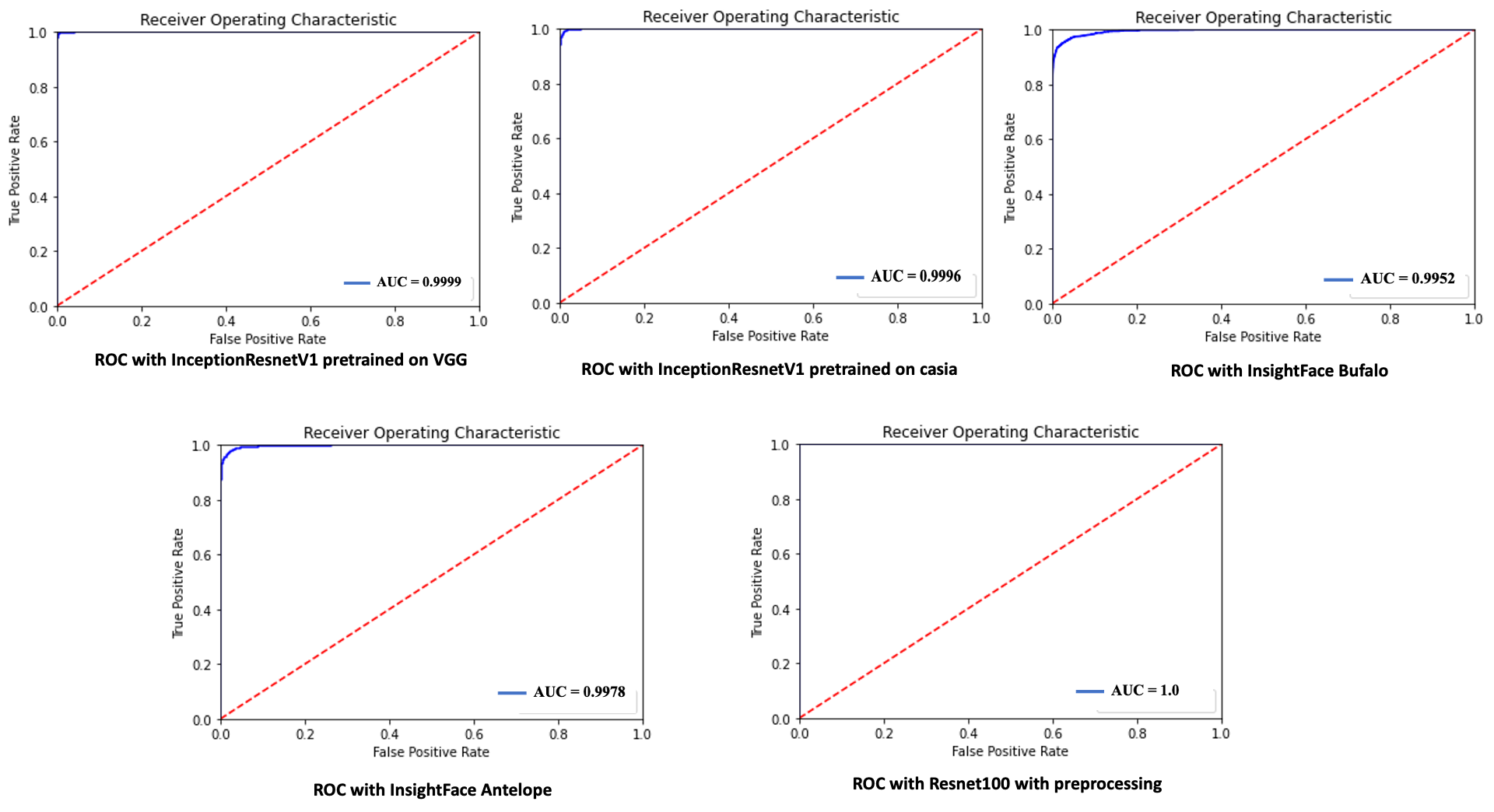}}
\caption{ROC Performance of Individual Classifiers Based on the AT\&T Face Dataset}
\label{att_single}
\end{figure*}

\begin{table}[htbp]
\caption{Individual Classifier AUC for Different Datasets}
\vspace{7pt} 
\begin{center}
\small
\begin{tabular}{|c|c|c|}
\hline
\textbf{\textit{Dataset used}} & \textbf{\textit{Classifier used}} & \textbf{\textit{AUC}} \\
\hline
AT\&T & InceptionResNetV1-VGG& .9999 \\
\hline
AT\&T & InceptionResNetV1-Casia& .9996 \\
\hline
AT\&T & InsightFace Buffalo& .9952 \\
\hline
AT\&T & InsightFace Antelope v2& .9978 \\
\hline
AT\&T & ResNet100 with preprocessing& 1.000 \\
\hline
GT & InceptionResNetV1-VGG& .9998 \\
\hline
GT & InceptionReNnetV1-Casia& .9995 \\
\hline
GT & InsightFace Buffalo& .9780 \\
\hline
GT & InsightFace Antelope v2& .9745 \\
\hline
GT & Resnet100 with preprocessing & 1.000 \\
\hline
LFW & InceptionResNetV1-VGG& .9904 \\
\hline
LFW & InceptionResNetV1-Casia& .9909 \\
\hline
LFW & InsightFace Buffalo& .9070 \\
\hline
LFW & InsightFace Antelope v2& .9175 \\
\hline
LFW & ResNet100 with preprocessing& .9701 \\
\hline
\end{tabular}
\label{ind_auc}
\end{center}
\end{table}

All five classifiers chosen for the secure hash pipeline performed exceptionally well on the AT\&T dataset, achieving an AUC exceeding 0.99 (shown in Table~\ref{ind_auc}). Similar success was observed on the GT dataset, where at least three classifiers surpassed an AUC of 0.99, and the remaining two achieved an AUC exceeding 0.97. The LFW dataset~\cite{LFWTech} yielded comparable results, with at least three classifiers again exceeding an AUC of 0.99. The designation `genuine' indicates a positive class, and `imposter' is marked as a negative class.

\subsection{End-to-end Benchmarking }

\begin{table}[htbp]
\small
\caption{End-to-end Benchmarking Performance}
\vspace{7pt} 
\centering
\begin{tabular}{|c|c|c|c|c|}
\hline
\textbf{\textit{Dataset}} & \textbf{\textit{Chaff points}} & \textbf{\textit{TPR}} & \textbf{\textit{TNR}} & \textbf{\textit{Classifier used}} \\
\hline
GT & 4000 & 81.20\% & 100.00\% & 5 \\
\hline
GT & 2000 & 84.63\% & 100.00\% & 5 \\
\hline
GT & 4000 & 91.46\% & 100.00\% & 4 \\
\hline
AT\&T & 4000 & 91.58\% & 100.00\% & 5 \\
\hline
AT\&T & 2000 & 93.27\% & 100.00\% & 5 \\
\hline
AT\&T & 4000 & 96.19\% & 100.00\% & 4 \\
\hline
\end{tabular}
\label{tab3}
\end{table}

As shown in Table~\ref{tab3}, the True Negative Rate (TNR) remained mostly at 100\% for the proposed framework when tested with both AT\&T and GT datasets. However, when the number of chaff points was increased from 2000 to 4000, the true positive rate slightly decreased. It was also observed that if the system had to satisfy only 4 out of all 5 classifiers, the true positive rate could improve even with more chaff points. Using a higher number of chaff points can increase the computation complexity, which in turn may make the system robust against  attack. Introducing additional classifiers strengthens the system by exponentially increasing the imposter's comparison workload ($(n+1)^C > n^C$), thus making the system further resilient against brute force attacks. Here, $C$ is the number of classifiers and $n$ is the number of chaff points. A combination of 4000 chaff points and 4~classifiers yielded the best TPR and TNR.

\begin{figure}[htbp]
\vspace{7pt} 
\centering
\vspace{-8pt}  
\includegraphics[width=0.7\columnwidth]{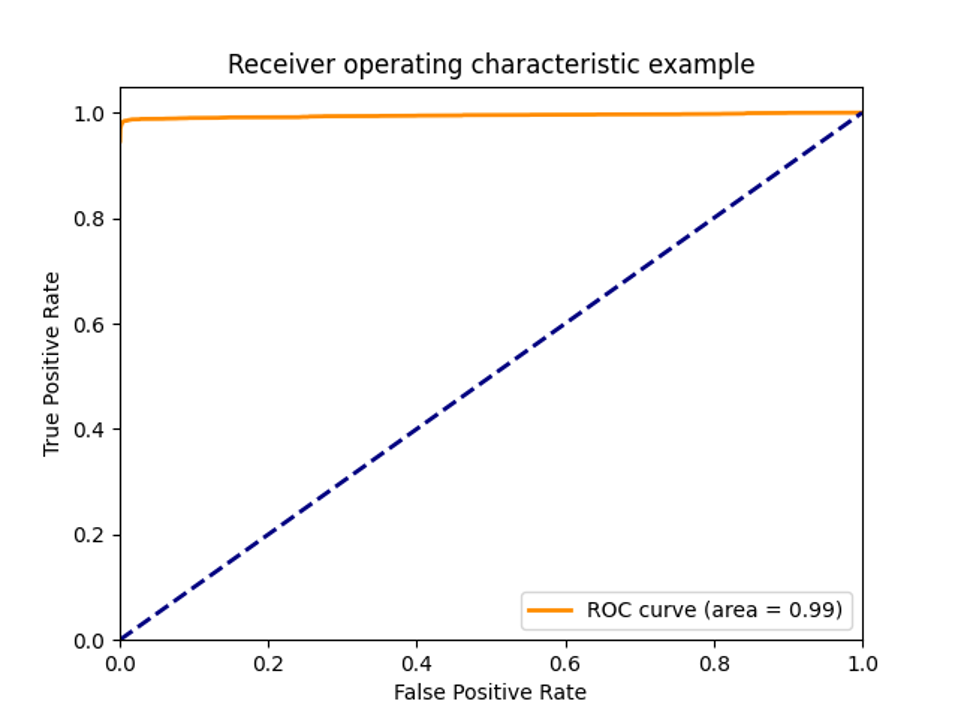}
\caption{End-to-end Performance Based on the AT\&T Face Dataset}
\label{roc_att}
\vspace{-8pt}  
\end{figure}

\begin{figure}[htbp]
\centering
\vspace{-8pt}  
\includegraphics[width=0.7\columnwidth]{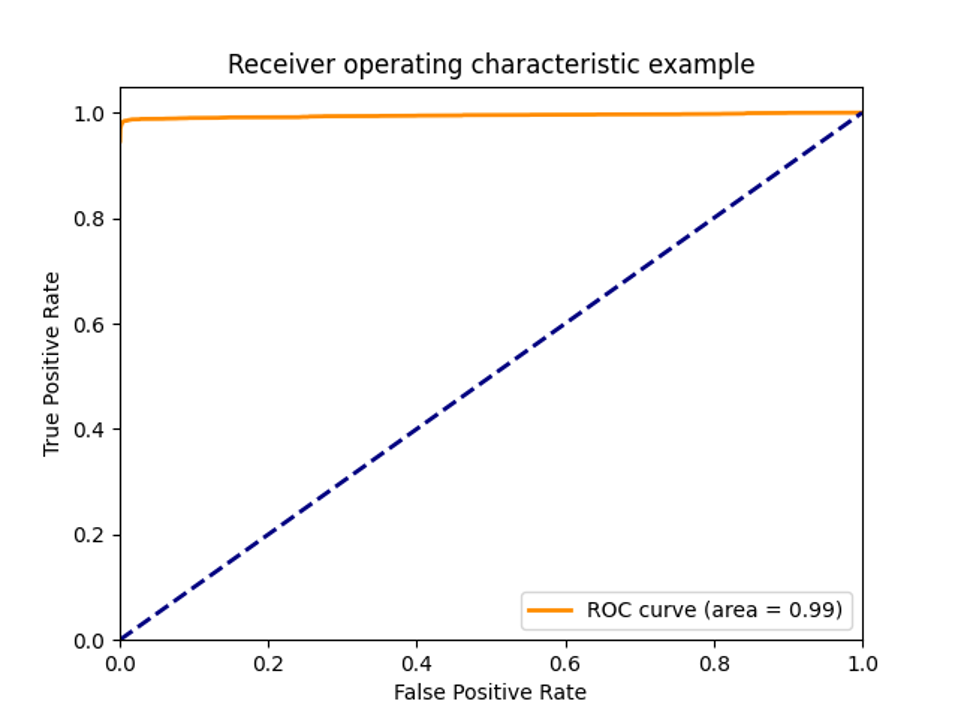}
\caption{End-to-end Performance Based on the GT Face Dataset}
\label{roc_gt}
\vspace{-6pt}  
\end{figure}

\begin{figure}[htbp]
\centering
\vspace{-6pt}  
\includegraphics[width=0.7\columnwidth, trim=2 2 2 2, clip]{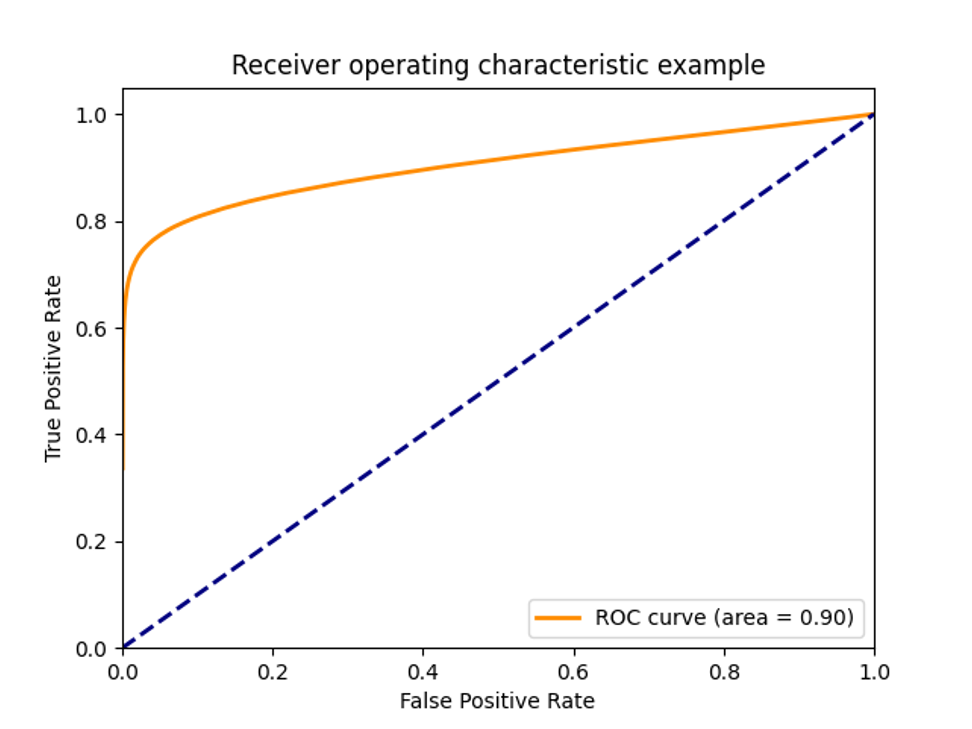}
\caption{End-to-end Performance Based on the LFW Face Dataset}
\label{roc_lfw}
\vspace{-8pt}  
\end{figure}

\subsection{ROC Analysis}

Figures~\ref{roc_att},~\ref{roc_gt}, and~\ref{roc_lfw} depict the study ROC curves for various datasets, all generated using the presented secure hash pipeline. Instead of evaluating individual classifiers, these ROC curves assess the performance of the complete biometric enrollment and authentication pipeline, which we refer to as the registration process. Within this pipeline, a sample is considered a true positive if a genuine person is successfully registered, and a false positive otherwise. The ROC curves (Figures~\ref{roc_att},~\ref{roc_gt},~\ref{roc_lfw}) depict the trade-off between security and accuracy in the registration process. These curves show how the rate of incorrectly accepting imposters (FPR) increases as we loosen the criteria for accepting genuine users (TPR). Imagine a vault containing 2001 images. The strictest system would only accept a single image with the highest cosine similarity score, while the most lenient would accept all 2001. Practically, accepting the top five highest-scoring images might be a good balance. This means the person can be authenticated if their true identity is among the top five closest matches out of the 2001 comparisons. As the ROC curves show, there's a point beyond which accepting more images doesn't significantly improve the chance of correctly identifying a genuine person, while increasing the risk of mistakenly accepting someone else.

\begin{table}[htbp]
\caption{End-to-end AUC Performance on Different Datasets}
\vspace{7pt} 
\centering
\small
\begin{tabular}{|c|c|}
\hline
\textbf{\textit{Dataset for benchmarking}} & \textbf{\textit{AUC}} \\
\hline
AT\&T face dataset & 0.9939 \\
\hline
Georgia Tech Face dataset & 0.9942 \\
\hline
LFW dataset & 0.9042 \\
\hline
\end{tabular}
\label{tab1}
\end{table}

As shown in \mbox{{Table}~\ref{tab1}}, the AT\&T and Georgia Tech Face datasets achieve an AUC greater than 0.99, while the LFW dataset, with its challenging variability in pose, expression, and lighting, yields a lower AUC of 0.9042.

\subsection{Time Complexity Analysis}

\begin{table}[htbp]
\caption{Time Complexity Performance}
\vspace{7pt} 
\centering
\small
\begin{tabular}{|c|c|}
\hline
\textbf{\textit{Event}} & \textbf{\textit{Time taken in seconds}} \\
\hline
Face detection and embedding & 1.15 \\
\hline
Matching (cosine similarity) & 0.31 \\
\hline
Secure hash generation & 0.01 \\
\hline
Complete end-to-end procedure & 1.47 \\
\hline
\end{tabular}
\label{timeline}
\end{table}

Our system, developed on a Windows laptop with an Intel i9 CPU and NVIDIA GeForce RTX 3080 GPU, demonstrates efficient performance. Face detection, preprocessing, and embedding creation take 1.15 seconds, matching 0.31 seconds, and secure hash generation 0.01 seconds(Table ~\ref{timeline}). The process completes in an average of 1.47 seconds. Comparable non-protected systems report 0.1 to 0.5 seconds for feature extraction and matching. Our system maintains competitive performance while introducing additional security performance.

\section{Conclusion}
In this paper, we proposed a cryptographic biometric template security protocol leveraging GenAI to enhance the security and privacy of face recognition. By harnessing GenAI's ability to generate highly realistic and diverse synthetic data, we demonstrated a $2^{\gamma}$ security against brute-force attacks. Our experimental results on the AT\&T, GT, and LFW datasets showed that our protocol preserves the average accuracy of the models used to generate the embeddings. We introduced a novel ROC curve computation method that evaluates our entire biometric pipeline, including multiple classifiers and GAN-generated chaff points. This comprehensive approach captures the synergistic effects of our multi-layered security system, unlike traditional ROC analyses that typically assess individual components separately. By utilizing GAN-generated synthetic data, we eliminated the risk of exposing real individuals' identities during training and verification, hence providing additional privacy protections. This work showcases GenAI's potential in constructing robust and secure end-to-end biometric systems. Future work may include exploring other GenAI methods to create more diverse samples and integrate them with other modalities, such as fingerprints, iris, and voice to develop multimodal biometric systems that are more resilient to attacks.

As GenAI continues its advance, one can envision a future where biometric security protocols continuously evolve to stay ahead of potential threats, enabling a new wave of secure and private biometric systems.

\section*{Acknowledgment}
This work was supported in part by a grant from CBL, an NSF IUCRC. Dr. Derakhshani is also a consultant for Jumio.

{\small
\bibliographystyle{ieee}
\bibliography{egbib}
}

\end{document}